\documentclass[runningheads]{llncs}
\usepackage[T1]{fontenc}
\usepackage{graphicx}
\usepackage{subcaption}

\usepackage{xcolor}
\definecolor{LightGray}{gray}{0.9}
\usepackage{hyperref}
\usepackage{listings}

\definecolor{codegreen}{rgb}{0,0.6,0}
\definecolor{codegray}{rgb}{0.5,0.5,0.5}
\definecolor{codepurple}{rgb}{0.58,0,0.82}
\definecolor{backcolour}{rgb}{0.95,0.95,0.92}

\lstdefinestyle{mystyle}{
    backgroundcolor=\color{backcolour},   
    commentstyle=\color{codegreen},
    keywordstyle=\color{magenta},
    numberstyle=\tiny\color{codegray},
    stringstyle=\color{codepurple},
    basicstyle=\ttfamily\footnotesize,
    breakatwhitespace=false,         
    breaklines=true,                 
    captionpos=b,                    
    keepspaces=true,                 
    numbers=left,                    
    numbersep=5pt,                  
    showspaces=false,                
    showstringspaces=false,
    showtabs=false,                  
    tabsize=2
}

\newcommand{\repeatthanks}{\textsuperscript{\thefootnote}}

\begin{document}
\title{A Retrospective of the Tutorial on Opportunities and Challenges of Online Deep Learning}
\titlerunning{Opportunities and Challenges of Online Deep Learning}
%
\author{Cedric~Kulbach\inst{1}\orcidID{0000-0002-9363-4728}\thanks{Equal Contribution.}
\and Lucas~Cazzonelli\inst{1}\orcidID{0000-0003-2886-1219}\repeatthanks
\and Hoang-Anh~Ngo\inst{2}\orcidID{0000-0002-7583-753X}\repeatthanks
\and Minh-Huong~Le-Nguyen\inst{3}\orcidID{0000-0002-5333-6785}
\and Albert~Bifet\inst{2,3}\orcidID{0000-0002-8339-7773}
}
\authorrunning{Kulbach C. et al.}
%
\institute{ 
 FZI Research Center for Information Technology, \\
  Haid-und-Neu-Str. 10-14, 76131 Karlsruhe, Germany \\
  \email{cedric.kulbach@gmail.com, cazzonelli@fzi.de} 
  \and AI Institute, University of Waikato, Hamilton, New Zealand  \\
  \email{abifet@waikato.ac.nz, h.a.ngo@sms.ed.ac.uk}
  \and LCTI, T\'el\'ecom Paris, Institut Polytechnique de Paris, France \\
  \email{minh.lenguyen@telecom-paris.fr}
  }
\maketitle          

\begin{abstract}
Machine learning algorithms have become indispensable in today's world. They support and accelerate the way we make decisions based on the data at hand. This acceleration means that data structures that were valid at one moment could no longer be valid in the future. With these changing data structures, it is necessary to adapt machine learning (ML) systems incrementally to the new data. This is done with the use of online learning or continuous ML technologies. While deep learning technologies have shown exceptional performance on predefined datasets, they have not been widely applied to online, streaming, and continuous learning. In this retrospective of our tutorial titled \textit{Opportunities and Challenges of Online Deep Learning} held at ECML PKDD 2023, we provide a brief overview of the opportunities but also the potential pitfalls for the application of neural networks in online learning environments using the frameworks \texttt{River} and \texttt{Deep-River}. 

\end{abstract}

\section{Introduction}

\begin{figure}[htbp]
\centerline{\includegraphics[scale=0.4]{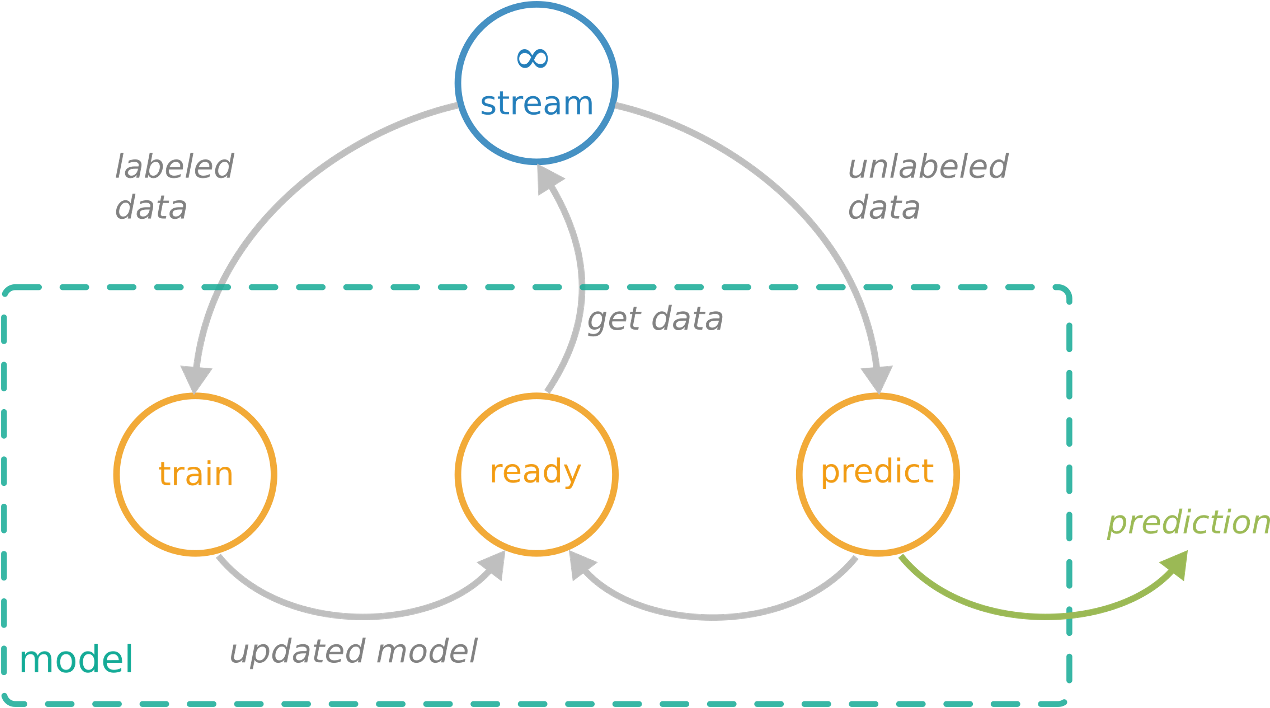}}
\caption{Structure of the interaction between data stream and prediction model (adapted from ~\cite{montiel2019}).}
\label{stream-structure}
\end{figure}

In many applications, machine learning models must deal with real-time data rather than self-contained data sets that are fully available before training. Therefore, it is critical to have access to algorithms that can process data that arrive continuously in the form of data streams. 
According to Bifet et al.~\cite{bifet2018}, such machine learning models operating in streaming environments must be able to:
\begin{enumerate}
    \label{rq:online_learning}
    \item Process an instance at a time, and inspect it (at most) once.
    \item Use a limited amount of time to process each instance.
    \item Use a limited amount of memory.
    \item Be ready to give an answer (e.g. prediction) at any time.
    \item Adapt to temporal changes.
\end{enumerate}
Figure~\ref{stream-structure} depicts how an online learning framework can comply with the data stream requirements \ref{rq:online_learning} for supervised learning tasks. 
The model processes labeled data points $\left(\overrightarrow{x},y\right)$ by updating the model while instead predicting a label $\hat{y}$ for each unlabeled instance $\overrightarrow{x}$. 
Thus, the model processes each instance from an evolving data stream, updates the underlying model, and is ready to predict at any time. 

Even now, the development of stream algorithms is still quite scattered and decentralized. Previously, algorithms were usually self-developed and maintained by the respective authors in various programming languages, with none of the existing frameworks being widely adopted within the online learning community. Currently, \texttt{River} is becoming not only a go-to library for online machine learning tasks but also a pioneer framework for the implementation of any new algorithm within the field.

A significant question in the context of the advancement of \texttt{River} is whether deep learning algorithms, which have been a staple in many batch learning frameworks for some time, can also fulfill the requirements \ref{rq:online_learning} and therefore be applied within online learning environments. To this end, we developed \texttt{Deep-River}, which combines the \texttt{River} API for online learning algorithms and \texttt{PyTorch} for the flexible development of neural architectures.

In our tutorial at ECML PKDD 2023, we gave an overview of the frameworks above and used them to explain the opportunities and challenges introduced by online deep learning. To this end, we demonstrated the transition from simple conventional machine learning models to sophisticated neural architectures while considering not only classification, regression, and anomaly detection metrics but also time and memory consumption which are key factors for the throughput of the underlying model.
In the following, we give a summary of these topics. 

\section{River and Fairness in Online Machine Learning}

\texttt{River}~\cite{JMLR:v22:20-1380} is the result of the merge of two other popular packages for stream learning in Python: \texttt{Creme}~\cite{creme2019} and \texttt{scikit-multiflow}~\cite{JMLR:v19:18-251}. \texttt{River} is not only a simple merge but also provides extra space for functionality, including support for mini-batches, improvements in processing time, incremental metrics for different methods, and more algorithms for each specific purpose. \texttt{River} is written mainly in Python, with certain core elements written in Cython~\cite{Behnel2011} for significant performance improvements. Currently, \texttt{River} not only supports numerous algorithms for a wide variety of purposes, including classification, regression, clustering, anomaly detection, representation learning, multi-label and multi-output learning but also provides utilities for other online tasks such as feature extraction and selection, pre-processing, progressive model validation and pipelines.

All models in \texttt{River} provide two core functions: learn (training/fitting) and predict. The learning process takes place via the either \texttt{learn\_one} or \texttt{learn\_many} method, while the predictions are made via the \texttt{predict\_one} (classification, regression, clustering), \texttt{predict\_proba\_one} (classification), or \texttt{score\_one} method depending on the learning task. Similarly, \texttt{Transformer} objects in this package transform an input via the \texttt{transform\_one} method.

To fully satisfy the requirements of online ML, which means that for instances to be processed once at a time, instead of \texttt{numpy.ndarray}, dictionaries are used as the default data structure in \texttt{River}~\cite{Gorelick2020-dw} since they provide the following advantages:

\begin{enumerate}
    \item Efficient storage of one-dimensional data with $O(1)$ lookup and insertion (under the assumption that samples are relatively small),
    \item Data access by name (key) rather than by position,
    \item Ability to store different types of data, and
    \item Flexibility in handling new features appearing within the data stream.
\end{enumerate}

Last but not least, \texttt{River} is designed to provide flexibility and ease of use in the deployment of stream learning in both academic and industrial settings. As such, \texttt{River} is gradually becoming the go-to library for machine learning for stream data, bringing together a large community of practitioners, enthusiasts, and researchers. 



For instance, BNP Paribas, a leading global banking group with approximately 33 million clients worldwide~\cite{ebf2019}, adopted \texttt{River} as part of its software architecture to tackle major use cases of stream learning within the banking industry~\cite{Barry2021}, including: 

\begin{enumerate}
    \item Real-time fraud or money laundering detection through suspicious transactions,
    \item Predictive maintenance of network IoT devices through stream anomaly detection of multiple and heterogeneous data streams, and
    \item Explainable credit risk scoring system for clients based on mining historical data to propose the most adapted, personalized loans plan, while staying in compliance with the European General Data Protection Regulation (GDPR).
\end{enumerate}


Within the tutorial, we presented several real-world use cases with \texttt{River}, including examples of classification, regression, anomaly detection, and clustering methods. Moreover, we have also presented a demonstration on how to conduct a real-time result visualization using other open-source libraries like \texttt{holoviews}~\cite{https://doi.org/10.5281/zenodo.596560}, \texttt{panel}~\cite{panel2023} and \texttt{streamz}~\cite{streamz2022} that enable the creation of interactive web-apps or dashboards and the build of pipelines to manage continuous streams of data seamless.

\subsection{Conventional Fairness-Aware Learning Methods}

Recently, fairness in machine learning has been gaining substantial attention, with the primary objective of identifying and removing bias and discrimination. For conventional batch machine learning approaches, methods within this field can be categorized into three main groups: pre-processing, in-processing, and post-processing methods.

\textbf{Pre-processing methods} work under the assumption that obtained data should already be discrimination-free. As such, these methods are \textit{model-agnostic}, which means that any machine learning model should be applicable after such a process. To obtain such a state, the following approaches are usually adopted:

\begin{itemize}
    \item \textbf{Massaging} \cite{Kamiran2009}: re-labeling certain instances close to the decision boundary to restore balance.
    \item \textbf{Re-weighting}~\cite{Calders2009}: assigning different weights to different groups to help benefit the deprived community.
    \item \textbf{Sampling}: including approaches that cannot directly work with weight. Two of the most well-known works within this section include:
    \begin{itemize}
        \item \textbf{Synthetic Minority Oversampling Techniques (SMOTE)}~\cite{10.5555/1622407.1622416} that generates synthetic observations of the minority class by interpolating between neighboring data points.
        \item A combination of \textbf{massaging} and \textbf{preferential sampling (PS)}~\cite{Kamiran2011} to obtain an effective decrease in discrimination with a minimal loss in accuracy.
    \end{itemize}
\end{itemize}

\textbf{In-processing methods} directly modify the algorithm to ensure it will generate fair outcomes. As such, contrary to pre-processing methods, the methods of this type are algorithm-specific. A popular approach of this type is to incorporate fairness constraints into the learner's objective function, either to create a joint loss function~\cite{pmlr-v81-dwork18a} or to reduce a fair classification problem to a series of cost-sensitive classification problems to yield the classifier with the lowest error subject to the desired fairness constraints~\cite{Padala2020,JMLR:v20:18-262,https://doi.org/10.48550/arxiv.1507.05259}.

\textbf{Post-processing methods} directly modify the obtained results of a trained model to ensure fairness and mitigate bias. Examples of such methods include:

\begin{itemize}
    \item Relabeling of leaves of a decision tree model~\cite{Kamiran2010},
    \item Correcting the confidence values of the classification rules based on a family of formal measures of discrimination~\cite{Pedreschi2009},
    \item Implementing the hypothesis that discrimination is usually seen close to the decision boundary via decision-theoretic concepts by exploiting the low confidence region (of a single/ensemble) and disagreement region (of an ensemble)~\cite{Kamiran2012}, or
    \item Shifting the decision boundary of the learner for the protected group based on the theory of margins for boosting while addressing the pitfalls of naive modification on biased algorithms~\cite{Fish2016}; 
\end{itemize}

\subsection{Fairness-Aware Stream Learning}

Several previous works have proposed adaptations of conventional fairness-aware, discrimination-mitigation methods to the stream learning environment. 

One of the first proposals within this field is a chunk-based intervention that modifies the input streaming data such that the outcome of any classifier should be discrimination-free~\cite{Iosifidis2019}. Concept drift is tackled by stream classifiers (Hoeffding Trees, Accuracy Updated Ensembles, and Naive Bayes), while fairness is ensured by either massaging or re-weighting upon each data chunk.

Fairness Aware Hoeffding Tree (FAHT)~\cite{Zhang2019}, an extension of the well-known Hoeffding Tree algorithm~\cite{Domingos2000} that effectively builds and updates decision trees with data streams, accounts for fairness by considering fairness gain along with the information in the splitting criterion of the decision tree. The idea of fairness gain is constructed to align with the idea of information gain, and for both, it holds that the higher the reduction, the better. These two concepts are combined to form Fair Information Gain (FIG). 

Online fairness and class imbalance-aware boosting (FABBOO)~\cite{Iosifidis2020} resolves the fairness problem by adjusting the decision boundary on demand through an online boosting classifier~\cite{https://doi.org/10.48550/arxiv.1206.6422}. The algorithm also considers and adapts to changes in the distribution of the data stream through time (concept drift) by employing Adaptive Hoeffding Trees~\cite{Bifet2009} as weak learners. Moreover, within the same paper, the concept of cumulative fairness, which monitors discriminatory outcomes until the current time point $t$ of the data stream, is introduced.

Continuous Synthetic Minority Oversampling Technique (C-SMOTE)~\cite{Bernardo2020}, inspired by the original SMOTE algorithm, resolves the problem of an imbalanced stream of data with the presence of concept drift by using ADWIN~\cite{Bifet2007} to save the most recently-observed data points and apply SMOTE to the minority class samples stored in such window. ADWIN uses the threshold $\delta$ to determine two error levels, warning and change. 

\begin{itemize}
    \item If the error exceeds the warning level, ADWIN assumes that a concert drift starts to occur, and it will start collecting new samples in a new window. 
    \item If the error exceeds the change level, ADWIN will assume that such drift has occurred and will replace the previously-existed window with the newly created one.
\end{itemize}

Most recently, a work presented at ECML PKDD 2023 under Journal Track proposes FAC-Fed~\cite{Badar2023}, which adapts federated learning to alleviate discrimination while detecting concept drift via Early Drift Detection Method (EDDM)~\cite{Manuel2006}. The discriminatory behavior of the model is measured using statistical parity and equal opportunity~\cite{Verma2018}. When the discrimination scores exceed a certain threshold $\epsilon$, the proposed continuous fairness-aware synthetic over-sampling technique (CFSOTE) is deployed.

\section{Online Deep Learning and Deep-River}

Deep learning models have set new performance standards in computer vision~\cite{redmonYouOnlyLook2016,krizhevskyImageNetClassificationDeep2017,radfordLearningTransferableVisual2021}, natural language processing~\cite{brownLanguageModelsAre2020,devlinBERTPretrainingDeep2019b} and countless other areas of application. 

Despite this success, deep learning models are still very rarely used in the realm of online learning. 
We see one of the reasons for the slow adoption of online deep learning in the currently very limited software support within the field. 
While a vast number of libraries are enabling the implementation, training, and deployment of deep learning architectures for batch learning problems, the development of tools for the same tasks in streaming environments has received little attention. 

With the \texttt{Deep-River} library, we aim to narrow this gap by providing the necessary tools for online learning researchers and practitioners to easily build and evaluate a wide range of neural network architectures. 

\subsection{Deep-River}

\texttt{Deep-River} is a Python library that provides an interface between the API of the \texttt{River} online learning framework and the deep learning framework \texttt{PyTorch}.
As such, it includes wrapper classes for the most commonly encountered online learning tasks, such as classification, regression, time-series forecasting and anomaly detection. 

As one of the most commonly used and most feature-rich deep learning frameworks \texttt{PyTorch}~\cite{paszkePyTorchImperativeStyle2019a} is a logical choice for an online deep learning backend. 
Compared to popular alternatives like \texttt{Tensorflow}~\cite{abadiTensorFlowSystemLargescale2016}, \texttt{PyTorch} offers a much more efficient conversion of Python dictionaries into Tensor objects required for automatic differentiation. 
It also enables a wide variety of architectures and the reuse of existing batch learning implementations that could hardly be matched by a possibly more efficient custom backend. 

\subsection{Example: Anomaly Detection with Deep-River}

As an example of the performance and ease of use of online deep learning with \texttt{Deep-River}, we perform anomaly detection on a data stream of credit card transactions~\cite{dalpozzoloLearnedLessonsCredit2014b}, some of which are fraudulent.

For this purpose, we use an autoencoder network, whose reconstruction error can be used as a score for the \textit{anomalousness} of individual data instances~\cite{sakuradaAnomalyDetectionUsing2014a}.  
We calculate anomaly scores for each sample in the data stream using both an autoencoder with a single hidden layer of 64 units and a Half Space Trees~\cite{tanFastAnomalyDetection2011a}, which is a commonly used conventional online anomaly detection approach. 
We implement the autoencoder network as a custom \texttt{PyTorch} module (see Python code in Appendix~\ref{code:anom_detection}) and wrap it using the \texttt{Autoencoder} class provided by \texttt{Deep-River}. 
To decide whether the current instance is anomalous, we label each instance whose score surpasses the 99th percentile of previous scores using Rivers \texttt{QuantileFilter}. 
\begin{figure}[hb]
\includegraphics[width=\textwidth]{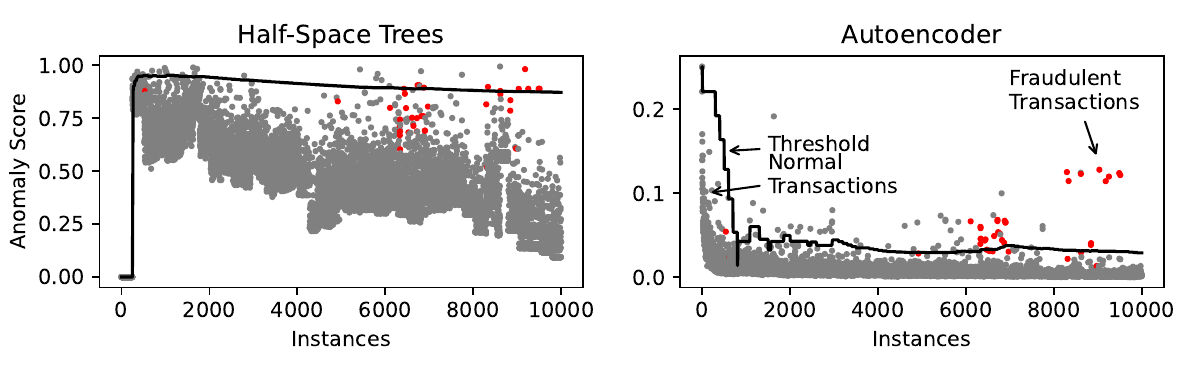}
\caption{Anomaly scores and decision boundaries of autoencoder- and Half-Space Trees anomaly detectors for stream consisting of credit card transactions~\cite{dalpozzoloLearnedLessonsCredit2014b}. Anomalies are shown in red.} \label{anom_scores}
\end{figure}
We then iterate over the data stream, first calculating the anomaly label based on the reconstruction error caused by the current sample and then adapting the model to it. 

The anomaly scores and decision boundaries resulting from this approach are visualized in Figure~\ref{anom_scores}. Here, it can be seen that the scores of anomalous transactions are much more separated from non-anomalous ones when using the autoencoder model compared to Half-Space Trees. Consequently, the autoencoder-based approach clearly outperforms the competing model with an F1 Score of 46.67\% compared to Half-Space Tree’s 22.73\%. 

\subsection{GPU-Usage in Online Deep Learning}

Due to the high degree of parallelism of most deep learning architectures, GPUs, which offer a large number of small processing units, are inherently well suited for running such models~\cite{ohGPUImplementationNeural2004}. 
This degree of parallelism is mainly driven by the dimensionality of the input data as well as the size of the network to be trained or used for inference. Due to the requirements for online learning
, online deep learning models must typically operate on individual examples instead of mini-batches of multiple instances, which reduces the effective dimensionality of model inputs significantly compared to the same model operating in a conventional batch learning environment. 

\begin{figure}[ht]
  \begin{subfigure}[b]{0.475\textwidth}
    \includegraphics[width=\linewidth]{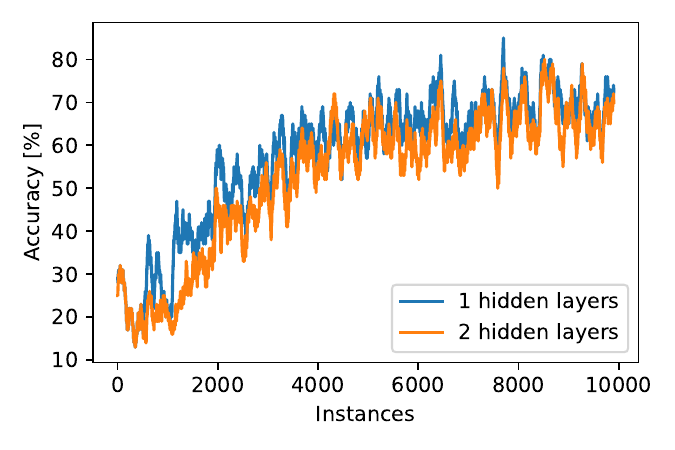}
    \caption{Prequential accuracy of MLPs with one or two hidden layers.}
    \label{subfig:accuracy_vs_n_hidden_layers}
  \end{subfigure}
  \begin{subfigure}[b]{0.475\textwidth}
    \includegraphics[width=\linewidth]{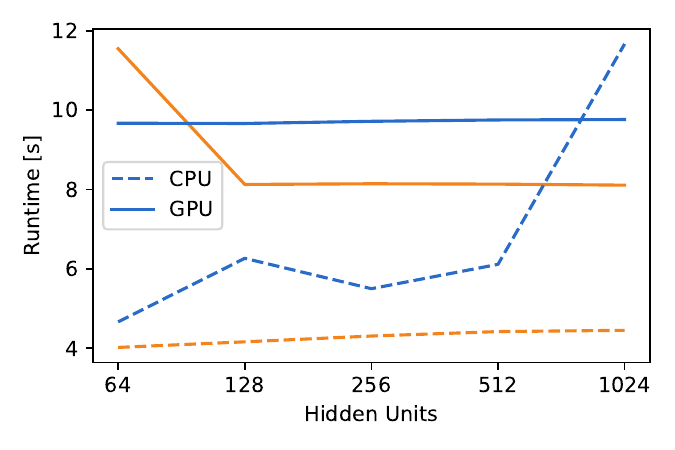}
    \caption{Runtime on an RTX 3090 GPU compared to an Intel i5-9600K CPU.}
    \label{subfig:network_size_gpu_vs_cpu}
  \end{subfigure}
  \caption{Results of prequential evaluations run with different MLP architectures on the first 10,000 samples of the \textit{Insects abrupt} classification dataset~\cite{souzaChallengesBenchmarkingStream2020d}.}
\end{figure}

Furthermore, in online deep learning, small networks are typically preferable over larger ones. This is, on one hand, due to the fact that online learning models must be able to predict as accurately as possible at any given time (see Requirements~\ref{rq:online_learning}) making larger architectures that are commonly used in batch learning (e.g., transformers) oftentimes unsuitable. An example of this can be seen in Figure~\ref{subfig:accuracy_vs_n_hidden_layers}, which shows the performance of multi-layer perceptrons with one or two layers. While the deeper network performs similarly to the single hidden layer model when given enough time to adapt to the current data concept, it yields significantly lower accuracy at the start of the stream as well as immediately prior to concept drifts. As a result, the smaller model is preferred when aiming to maximize the mean accuracy of the predictor. 

Additionally, online learning models are often run on hardware with limited computational capabilities, such as IoT devices, smartphones, etc.~\cite{shanthamalluBriefSurveyMachine2017}, which severely restricts the model size. 
All in all, neural networks operating in online learning environments, therefore, typically feature a much lower degree of parallelism than in batch learning, meaning that the usage of GPUs in this area is, therefore, rarely advantageous over CPU-based execution. 

This is also reflected in Figure~\ref{subfig:network_size_gpu_vs_cpu}, which shows the total runtime of MLPs with one or two hidden layers in relation to the number of units when evaluated on the first 10 thousand samples of the Insects abrupt data stream~\cite{souzaChallengesBenchmarkingStream2020d}. While the runtime difference between GPU and CPU execution decreases with an increasing number of hidden units, the runtimes achieved with the CPU are significantly lower for all but the largest network with two hidden layers consisting of 1024. As previously discussed, networks of this size are unnecessarily large due to their slower convergence compared to lighter architectures for most typical online learning tasks. Generally speaking, the usefulness of GPUs for deep online learning is thus relatively limited. 
Exceptions to this could be high-dimensional, more complex data such as image or video streams, which would benefit from larger neural architectures to a much higher degree than the typically relatively simple tasks encountered in online learning. However, to ensure adequate performance, such data-hungry models would likely have to be trained prior to execution on a data stream, which would technically violate the online learning requirements (see Requirements~\ref{rq:online_learning}).

\section{Conclusion}

In this brief retrospective of our tutorial on \textit{Opportunities and Challenges in Online Deep Learning} presented at ECML PKDD 2023, we gave a brief introduction to online learning and its differences compared to conventional machine learning and examined the design principles and advantages of the online learning framework \texttt{River}. 
We also provided a short overview as to why fairness is an important aspect of machine learning and how fairness can be induced in online learning. 
In the second part of this paper, we outlined the potential of deep learning models to set new standards for performance in online learning applications and introduced the \texttt{Deep-River} framework, which builds on top of \texttt{River} and \texttt{PyTorch} to facilitate the development and evaluation of online deep learning models.
To demonstrate the capabilities of \texttt{Deep-River} and online deep learning, we implemented an online anomaly detector based on an autoencoder network, which clearly outperformed a conventional anomaly detection approach. 
Lastly, we explained the limited usefulness of GPUs in online deep learning caused by its typically much lower degree of parallelism as an example of one of the challenges of the field. 

All material related to the tutorial, including presentation slides and practical demonstration with \texttt{Jupyter notebooks}, can be found within the associated \href{https://github.com/lucasczz/deep-river-demo-23}{Github repository}.

%
%

\bibliographystyle{splncs04}
\bibliography{bibliography}

\begin{thebibliography}{10}
\providecommand{\url}[1]{\texttt{#1}}
\providecommand{\urlprefix}{URL }
\providecommand{\doi}[1]{https://doi.org/#1}

\bibitem{streamz2022}
streamz 0.6.4 (Jul 2022), \url{https://pypi.org/project/streamz/}

\bibitem{panel2023}
Panel: The powerful data exploration \& web app framework for python (Sep
  2023), \url{https://pypi.org/project/panel/}

\bibitem{abadiTensorFlowSystemLargescale2016}
Abadi, M., Barham, P., Chen, J., Chen, Z., Davis, A., Dean, J., Devin, M.,
  Ghemawat, S., Irving, G., Isard, M., Kudlur, M., Levenberg, J., Monga, R.,
  Moore, S., Murray, D.G., Steiner, B., Tucker, P., Vasudevan, V., Warden, P.,
  Wicke, M., Yu, Y., Zheng, X.: {{TensorFlow}}: {{A}} system for large-scale
  machine learning (May 2016). \doi{10.48550/arXiv.1605.08695}

\bibitem{Badar2023}
Badar, M., Nejdl, W., Fisichella, M.: {FAC}-fed: Federated adaptation for
  fairness and concept drift aware stream classification. Machine Learning
  \textbf{112}(8),  2761--2786 (Jul 2023). \doi{10.1007/s10994-023-06360-7},
  \url{https://doi.org/10.1007/s10994-023-06360-7}

\bibitem{Manuel2006}
Baena-García, M., Campo-Ávila, J., Fidalgo-Merino, R., Bifet, A., Gavald, R.,
  Morales-Bueno, R.: Early drift detection method. In: Fourth International
  Workshop on Knowledge Dsicovery from Data Streams. vol.~6, pp. 77--86 (9
  2006)

\bibitem{Barry2021}
Barry, M., Bifet, A., Chiky, R., Montiel, J., Tran, V.T.: Challenges of machine
  learning for data streams in the banking industry. In: Big Data Analytics,
  pp. 106--118. Springer International Publishing (2021).
  \doi{10.1007/978-3-030-93620-4_9}

\bibitem{Behnel2011}
Behnel, S., Bradshaw, R., Citro, C., Dalcin, L., Seljebotn, D.S., Smith, K.:
  Cython: The best of both worlds. Computing in Science \& Engineering
  \textbf{13}(2),  31--39 (Mar 2011). \doi{10.1109/mcse.2010.118},
  \url{https://doi.org/10.1109/mcse.2010.118}

\bibitem{Bernardo2020}
Bernardo, A., Gomes, H.M., Montiel, J., Pfahringer, B., Bifet, A., Valle, E.D.:
  C-{SMOTE}: Continuous synthetic minority oversampling for evolving data
  streams. In: 2020 {IEEE} International Conference on Big Data (Big Data).
  {IEEE} (Dec 2020). \doi{10.1109/bigdata50022.2020.9377768},
  \url{https://doi.org/10.1109/bigdata50022.2020.9377768}

\bibitem{Bifet2007}
Bifet, A., Gavald{\`{a}}, R.: Learning from time-changing data with adaptive
  windowing. In: Proceedings of the 2007 {SIAM} International Conference on
  Data Mining. Society for Industrial and Applied Mathematics (Apr 2007).
  \doi{10.1137/1.9781611972771.42},
  \url{https://doi.org/10.1137/1.9781611972771.42}

\bibitem{Bifet2009}
Bifet, A., Gavald{\`{a}}, R.: Adaptive learning from evolving data streams. In:
  Advances in Intelligent Data Analysis {VIII}, pp. 249--260. Springer Berlin
  Heidelberg (2009). \doi{10.1007/978-3-642-03915-7_22}

\bibitem{bifet2018}
Bifet, A., Gavald{\`a}, R., Holmes, G., Pfahringer, B.: Machine Learning for
  Data Streams with Practical Examples in {{MOA}}. {MIT Press} (2018)

\bibitem{brownLanguageModelsAre2020}
Brown, T.B., Mann, B., Ryder, N., Subbiah, M., Kaplan, J., Dhariwal, P.,
  Neelakantan, A., Shyam, P., Sastry, G., Askell, A., Agarwal, S.,
  {Herbert-Voss}, A., Krueger, G., Henighan, T., Child, R., Ramesh, A.,
  Ziegler, D.M., Wu, J., Winter, C., Hesse, C., Chen, M., Sigler, E., Litwin,
  M., Gray, S., Chess, B., Clark, J., Berner, C., McCandlish, S., Radford, A.,
  Sutskever, I., Amodei, D.: Language {{Models}} are {{Few-Shot Learners}} (Jul
  2020). \doi{10.48550/arXiv.2005.14165}

\bibitem{Calders2009}
Calders, T., Kamiran, F., Pechenizkiy, M.: Building classifiers with
  independency constraints. In: 2009 {IEEE} International Conference on Data
  Mining Workshops. {IEEE} (Dec 2009). \doi{10.1109/icdmw.2009.83},
  \url{https://doi.org/10.1109/icdmw.2009.83}

\bibitem{10.5555/1622407.1622416}
Chawla, N.V., Bowyer, K.W., Hall, L.O., Kegelmeyer, W.P.: Smote: Synthetic
  minority over-sampling technique. J. Artif. Int. Res.  \textbf{16}(1),
  321–357 (jun 2002)

\bibitem{https://doi.org/10.48550/arxiv.1206.6422}
Chen, S.T., Lin, H.T., Lu, C.J.: An online boosting algorithm with theoretical
  justifications (2012). \doi{10.48550/ARXIV.1206.6422},
  \url{https://arxiv.org/abs/1206.6422}

\bibitem{dalpozzoloLearnedLessonsCredit2014b}
Dal~Pozzolo, A., Caelen, O., Le~Borgne, Y.A., Waterschoot, S., Bontempi, G.:
  Learned lessons in credit card fraud detection from a practitioner
  perspective. Expert Systems with Applications  \textbf{41}(10),  4915--4928
  (Aug 2014). \doi{10.1016/j.eswa.2014.02.026}

\bibitem{devlinBERTPretrainingDeep2019b}
Devlin, J., Chang, M.W., Lee, K., Toutanova, K.: {{BERT}}: {{Pre-training}} of
  {{Deep Bidirectional Transformers}} for {{Language Understanding}} (May
  2019). \doi{10.48550/arXiv.1810.04805}

\bibitem{Domingos2000}
Domingos, P., Hulten, G.: Mining high-speed data streams. In: Proceedings of
  the sixth {ACM} {SIGKDD} international conference on Knowledge discovery and
  data mining. {ACM} (Aug 2000). \doi{10.1145/347090.347107},
  \url{https://doi.org/10.1145/347090.347107}

\bibitem{pmlr-v81-dwork18a}
Dwork, C., Immorlica, N., Kalai, A.T., Leiserson, M.: Decoupled classifiers for
  group-fair and efficient machine learning. In: Friedler, S.A., Wilson, C.
  (eds.) Proceedings of the 1st Conference on Fairness, Accountability and
  Transparency. Proceedings of Machine Learning Research, vol.~81, pp.
  119--133. PMLR (23--24 Feb 2018),
  \url{https://proceedings.mlr.press/v81/dwork18a.html}

\bibitem{ebf2019}
{European Banking Federation}: Ebf position paper on ai in the banking industry
  (Jul 2019),
  \url{https://www.ebf.eu/wp-content/uploads/2020/03/EBF\_037419-Artificial-Intelligence-in-the-banking-sector-EBF.pdf}

\bibitem{Fish2016}
Fish, B., Kun, J., Lelkes, {\'{A}}.D.: A confidence-based approach for
  balancing fairness and accuracy. In: Proceedings of the 2016 {SIAM}
  International Conference on Data Mining. Society for Industrial and Applied
  Mathematics (Jun 2016). \doi{10.1137/1.9781611974348.17},
  \url{https://doi.org/10.1137/1.9781611974348.17}

\bibitem{Gorelick2020-dw}
Gorelick, M., Ozsvald, I.: High performance python. O'Reilly Media, Sebastopol,
  CA, 2 edn. (May 2020)

\bibitem{creme2019}
Halford, M., Bolmier, G., Sourty, R., Vaysse, R., Zouitine, A.: {creme}, a
  {P}ython library for online machine learning (2019),
  \url{https://github.com/MaxHalford/creme}

\bibitem{Iosifidis2020}
Iosifidis, V., Ntoutsi, E.: {\textdollar}{\textdollar}{\textbackslash}mathsf
  $\lbrace${FABBOO}$\rbrace${\textdollar}{\textdollar} - online fairness-aware
  learning under class imbalance. In: Discovery Science, pp. 159--174. Springer
  International Publishing (2020). \doi{10.1007/978-3-030-61527-7_11}

\bibitem{Iosifidis2019}
Iosifidis, V., Tran, T.N.H., Ntoutsi, E.: Fairness-enhancing interventions in
  stream classification. In: Lecture Notes in Computer Science, pp. 261--276.
  Springer International Publishing (2019). \doi{10.1007/978-3-030-27615-7_20}

\bibitem{Kamiran2009}
Kamiran, F., Calders, T.: Classifying without discriminating. In: 2009 2nd
  International Conference on Computer, Control and Communication. {IEEE} (Feb
  2009). \doi{10.1109/ic4.2009.4909197},
  \url{https://doi.org/10.1109/ic4.2009.4909197}

\bibitem{Kamiran2011}
Kamiran, F., Calders, T.: Data preprocessing techniques for classification
  without discrimination. Knowledge and Information Systems  \textbf{33}(1),
  1--33 (Dec 2011). \doi{10.1007/s10115-011-0463-8},
  \url{https://doi.org/10.1007/s10115-011-0463-8}

\bibitem{Kamiran2010}
Kamiran, F., Calders, T., Pechenizkiy, M.: Discrimination aware decision tree
  learning. In: 2010 {IEEE} International Conference on Data Mining. {IEEE}
  (Dec 2010). \doi{10.1109/icdm.2010.50},
  \url{https://doi.org/10.1109/icdm.2010.50}

\bibitem{Kamiran2012}
Kamiran, F., Karim, A., Zhang, X.: Decision theory for discrimination-aware
  classification. In: 2012 {IEEE} 12th International Conference on Data Mining.
  {IEEE} (Dec 2012). \doi{10.1109/icdm.2012.45},
  \url{https://doi.org/10.1109/icdm.2012.45}

\bibitem{krizhevskyImageNetClassificationDeep2017}
Krizhevsky, A., Sutskever, I., Hinton, G.E.: {{ImageNet}} classification with
  deep convolutional neural networks. Communications of the ACM
  \textbf{60}(6),  84--90 (May 2017). \doi{10.1145/3065386}

\bibitem{montiel2019}
L{\'o}pez, J.M.: Fast and Slow Machine Learning. Phd thesis, Télécom
  Paristech, Université Paris-Saclay, Paris, France (March 2019), available at
  \url{https://www.theses.fr/2019SACLT014.pdf}

\bibitem{JMLR:v22:20-1380}
Montiel, J., Halford, M., Mastelini, S.M., Bolmier, G., Sourty, R., Vaysse, R.,
  Zouitine, A., Gomes, H.M., Read, J., Abdessalem, T., Bifet, A.: River:
  machine learning for streaming data in python. Journal of Machine Learning
  Research  \textbf{22}(110), ~1--8 (2021),
  \url{http://jmlr.org/papers/v22/20-1380.html}

\bibitem{JMLR:v19:18-251}
Montiel, J., Read, J., Bifet, A., Abdessalem, T.: Scikit-multiflow: A
  multi-output streaming framework. Journal of Machine Learning Research
  \textbf{19}(72), ~1--5 (2018), \url{http://jmlr.org/papers/v19/18-251.html}

\bibitem{ohGPUImplementationNeural2004}
Oh, K.S., Jung, K.: {{GPU}} implementation of neural networks. Pattern
  Recognition  \textbf{37}(6),  1311--1314 (Jun 2004).
  \doi{10.1016/j.patcog.2004.01.013}

\bibitem{Padala2020}
Padala, M., Gujar, S.: {FNNC}: Achieving fairness through neural networks. In:
  Proceedings of the Twenty-Ninth International Joint Conference on Artificial
  Intelligence. International Joint Conferences on Artificial Intelligence
  Organization (Jul 2020). \doi{10.24963/ijcai.2020/315},
  \url{https://doi.org/10.24963/ijcai.2020/315}

\bibitem{paszkePyTorchImperativeStyle2019a}
Paszke, A., Gross, S., Massa, F., Lerer, A., Bradbury, J., Chanan, G., Killeen,
  T., Lin, Z., Gimelshein, N., Antiga, L., Desmaison, A., K{\"o}pf, A., Yang,
  E., DeVito, Z., Raison, M., Tejani, A., Chilamkurthy, S., Steiner, B., Fang,
  L., Bai, J., Chintala, S.: {{PyTorch}}: {{An Imperative Style}},
  {{High-Performance Deep Learning Library}} (Dec 2019).
  \doi{10.48550/arXiv.1912.01703}

\bibitem{Pedreschi2009}
Pedreschi, D., Ruggieri, S., Turini, F.: Measuring discrimination in
  socially-sensitive decision records. In: Proceedings of the 2009 {SIAM}
  International Conference on Data Mining. Society for Industrial and Applied
  Mathematics (Apr 2009). \doi{10.1137/1.9781611972795.50},
  \url{https://doi.org/10.1137/1.9781611972795.50}

\bibitem{radfordLearningTransferableVisual2021}
Radford, A., Kim, J.W., Hallacy, C., Ramesh, A., Goh, G., Agarwal, S., Sastry,
  G., Askell, A., Mishkin, P., Clark, J., Krueger, G., Sutskever, I.: Learning
  {{Transferable Visual Models From Natural Language Supervision}} (Feb 2021)

\bibitem{redmonYouOnlyLook2016}
Redmon, J., Divvala, S., Girshick, R., Farhadi, A.: You {{Only Look Once}}:
  {{Unified}}, {{Real-Time Object Detection}} (May 2016).
  \doi{10.48550/arXiv.1506.02640}

\bibitem{https://doi.org/10.5281/zenodo.596560}
Rudiger, P., Stevens, J.L., Bednar, J.A., Hansen, S.H., Liquet, M., {, Andrew},
  Nijholt, B., Mease, J., B, C., Randelhoff, A., Tenner, V., {Maxalbert},
  Kaiser, M., {Ea42gh}, {Stonebig}, Samuels, J., Raillard, D., Pevey, K., LB,
  F., Thomas, I., Madsen, M.S., Roelants, P., Tolmie, A., Stephan, D.,
  {Demetris Roumis}, Bois, J.: holoviz/holoviews: Version 1.17.1 (2023).
  \doi{10.5281/ZENODO.596560}, \url{https://zenodo.org/record/596560}

\bibitem{sakuradaAnomalyDetectionUsing2014a}
Sakurada, M., Yairi, T.: Anomaly {{Detection Using Autoencoders}} with
  {{Nonlinear Dimensionality Reduction}}. In: Proceedings of the {{MLSDA}} 2014
  2nd {{Workshop}} on {{Machine Learning}} for {{Sensory Data Analysis}}. pp.
  4--11. {{MLSDA}}'14, {Association for Computing Machinery}, {New York, NY,
  USA} (Dec 2014). \doi{10.1145/2689746.2689747}

\bibitem{shanthamalluBriefSurveyMachine2017}
Shanthamallu, U.S., Spanias, A., Tepedelenlioglu, C., Stanley, M.: A brief
  survey of machine learning methods and their sensor and {{IoT}} applications.
  In: 2017 8th {{International Conference}} on {{Information}},
  {{Intelligence}}, {{Systems}} \& {{Applications}} ({{IISA}}). pp.~1--8.
  {IEEE}, {Larnaca} (Aug 2017). \doi{10.1109/IISA.2017.8316459}

\bibitem{souzaChallengesBenchmarkingStream2020d}
Souza, V.M.A., dos Reis, D.M., Maletzke, A.G., Batista, G.E.A.P.A.: Challenges
  in {{Benchmarking Stream Learning Algorithms}} with {{Real-world Data}}. Data
  Mining and Knowledge Discovery  \textbf{34}(6),  1805--1858 (Nov 2020).
  \doi{10.1007/s10618-020-00698-5}

\bibitem{tanFastAnomalyDetection2011a}
Tan, S.C., Ting, K.M., Liu, T.F.: Fast {{Anomaly Detection}} for {{Streaming
  Data}}. In: Proceedings of the {{Twenty-Second International Joint
  Conference}} on {{Artificial Intelligence}} (2011)

\bibitem{Verma2018}
Verma, S., Rubin, J.: Fairness definitions explained. In: Proceedings of the
  International Workshop on Software Fairness. {ACM} (May 2018).
  \doi{10.1145/3194770.3194776}, \url{https://doi.org/10.1145/3194770.3194776}

\bibitem{JMLR:v20:18-262}
Zafar, M.B., Valera, I., Gomez-Rodriguez, M., Gummadi, K.P.: Fairness
  constraints: A flexible approach for fair classification. Journal of Machine
  Learning Research  \textbf{20}(75),  1--42 (2019),
  \url{http://jmlr.org/papers/v20/18-262.html}

\bibitem{https://doi.org/10.48550/arxiv.1507.05259}
Zafar, M.B., Valera, I., Rodriguez, M.G., Gummadi, K.P.: Fairness constraints:
  Mechanisms for fair classification (2015). \doi{10.48550/ARXIV.1507.05259},
  \url{https://arxiv.org/abs/1507.05259}

\bibitem{Zhang2019}
Zhang, W., Ntoutsi, E.: {FAHT}: An adaptive fairness-aware decision tree
  classifier. In: Proceedings of the Twenty-Eighth International Joint
  Conference on Artificial Intelligence. International Joint Conferences on
  Artificial Intelligence Organization (Aug 2019).
  \doi{10.24963/ijcai.2019/205}, \url{https://doi.org/10.24963/ijcai.2019/205}

\end{thebibliography}

\newpage
\appendix

\section{Code Example: Anomaly Detection}

\begin{lstlisting}[language=Python, caption={Code example for anomaly detection with \texttt{Deep-River}.}, label={code:anom_detection}]
import torch
from torch import nn
from deep_river.anomaly import Autoencoder, RollingAutoencoder

# Define PyTorch module 
class MyAutoencoder(nn.Module):
    def __init__(self, n_features, latent_dim=64):
        super().__init__()
        self.encoder = nn.Linear(n_features, latent_dim) 
        self.decoder = nn.Linear(latent_dim, n_features)

    def forward(self, X, **kwargs):
        z = torch.relu(self.encoder(X))
        return torch.sigmoid(self.decoder(z))

# Create anomaly detection pipeline
model = compose.Pipeline(
    preprocessing.MinMaxScaler(),
    anomaly.QuantileFilter(Autoencoder(module=MyAutoencoder, lr=0.25), q=0.99),
)
thresholder = model['QuantileFilter']

tracker = StreamTracker()

# Iterate over data stream
for x, y in tqdm(data):
    # Compute anomaly score and learn current instance
    score = model.score_one(x)
    model.learn_one(x)

    # Get anomaly label based on score
    pred = thresholder.classify(score)
    
    tracker.update(pred, score, thresholder.quantile.get())
\end{lstlisting}

\section{Demo: Live Visualization of Streaming Classification Metrics}

\begin{figure}
\centering
  \begin{subfigure}[b]{\textwidth}
    \includegraphics[width=\linewidth]{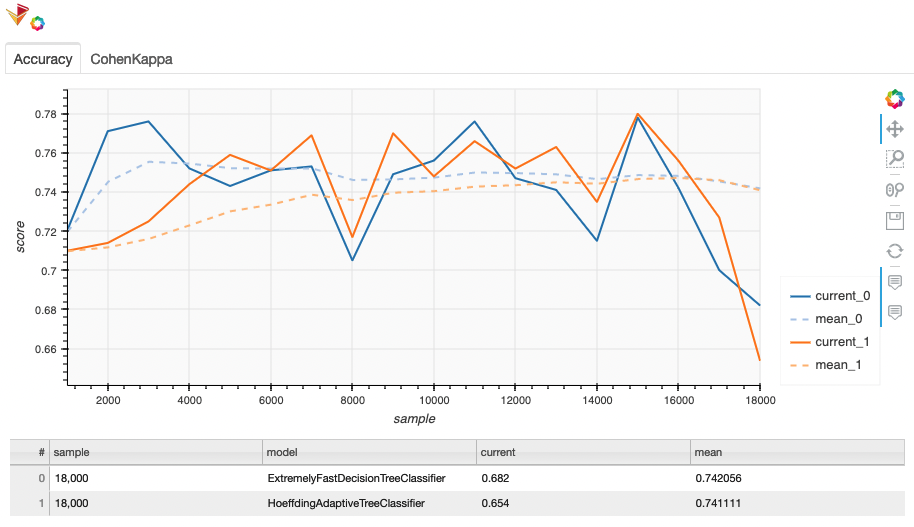}
    \label{subfig:layout-tabs-river-demo-accuracy}
    \caption{Results measured with accuracy metric.}
  \end{subfigure}
  \begin{subfigure}[b]{\textwidth}
    \includegraphics[width=\linewidth]{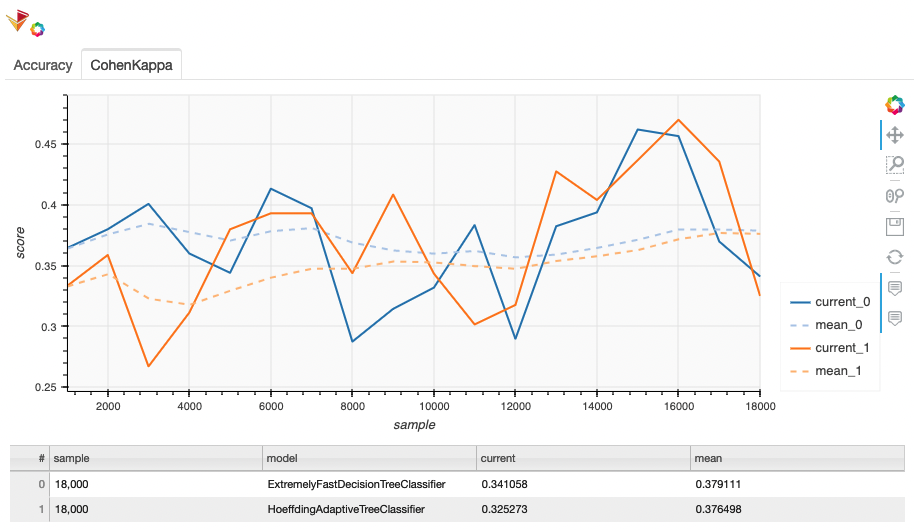}

    \label{subfig:layout-tabs-river-demo-cohenkappa}
    \caption{Results measured with Cohen-Kappa metric.}
  \end{subfigure}
  \caption{Capture of live visualization demo of \texttt{ExtremelyFastDecisionTreeClassifier} and \texttt{Hoeffding Tree Classifier} with Accuracy and Cohen-Kappa metrics.}
\end{figure}

\end{document}